\documentclass[10pt,twocolumn,letterpaper]{article}

\usepackage{wacv}              

\usepackage{graphicx}
\usepackage{amsmath}
\usepackage{amssymb}
\usepackage{booktabs}
\usepackage[accsupp]{axessibility}

\usepackage[pagebackref,breaklinks,colorlinks]{hyperref}

\usepackage[capitalize]{cleveref}
\crefname{section}{Sec.}{Secs.}
\Crefname{section}{Section}{Sections}
\Crefname{table}{Table}{Tables}
\crefname{table}{Tab.}{Tabs.}

\def\wacvPaperID{1069} 
\def\confName{WACV}
\def\confYear{2025}

\title{Distilling Aggregated Knowledge for Weakly-Supervised Video Anomaly Detection}

\author{Jash Dalvi\\
K J Somaiya Institute of Technology\\
{\tt\small jashdalvi99@gmail.com}
\and
Ali Dabouei \thanks{Corresponding authors.}\\
Carnegie Mellon University\\
{\tt\small ali.dabouei@gmail.com}
\and
Gunjan Dhanuka \\
Carnegie Mellon University\\
{\tt\small gdhanuka@cs.cmu.edu}
\and
Min Xu \protect\footnotemark[1]\\
Carnegie Mellon University\\
{\tt\small mxu1@cs.cmu.edu}
}

\begin{document}
\maketitle

\def \v{\mathbf{v}}
\def \z{\mathbf{z}}
\def \h{\mathbf{h}}

\begin{abstract}
Video anomaly detection aims to develop automated models capable of identifying abnormal events in surveillance videos. The benchmark setup for this task is extremely challenging due to: i) the limited size of the training sets, ii) weak supervision provided in terms of video-level labels, and iii) intrinsic class imbalance induced by the scarcity of abnormal events. In this work, we show that distilling knowledge from aggregated representations of multiple backbones into a single-backbone Student model achieves state-of-the-art performance. In particular, we develop a bi-level distillation approach along with a novel disentangled cross-attention-based feature aggregation network. Our proposed approach, \textbf{DAKD} (\textbf{D}istilling \textbf{A}ggregated \textbf{K}nowledge with \textbf{D}isentangled Attention), demonstrates superior performance compared to existing methods across multiple benchmark datasets. Notably, we achieve significant improvements of 1.36\%, 0.78\%, and 7.02\% on the UCF-Crime, ShanghaiTech, and XD-Violence datasets, respectively.

\end{abstract}

\section{Introduction}

Video Anomaly Detection (VAD) is a realization of automation based on video data which addresses the exhaustive labor and time requirements of video surveillance. The goal of a practical VAD system is to identify an activity that deviates from normal activities characterized by the training distribution \cite{sultani2018real}.

\begin{figure}
\centering
    \includegraphics[width=0.45\textwidth]{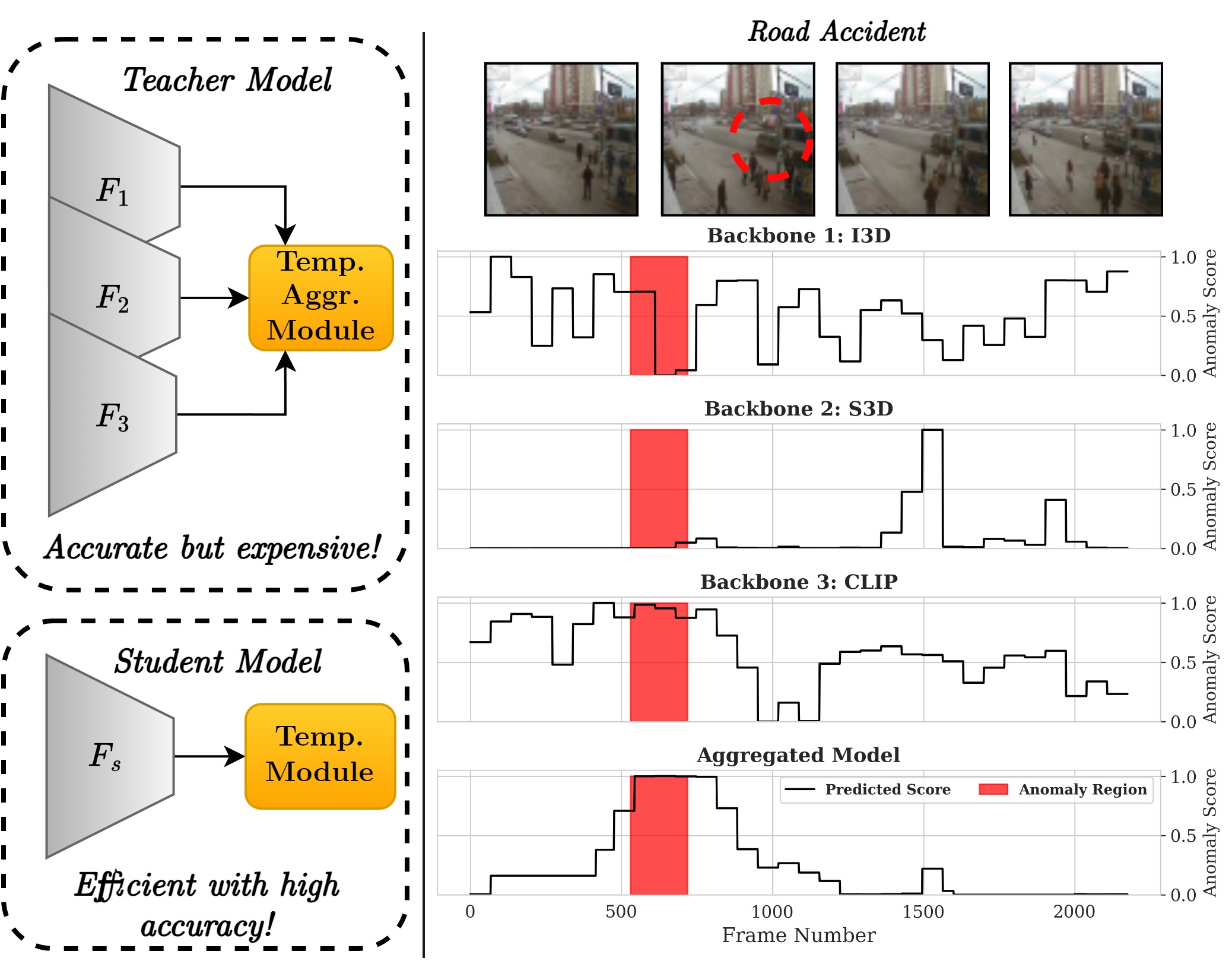}
    \caption{Left: A brief overview of our approach that distills the multi-backbone Teacher model's knowledge to the Student model. In the Teacher model, representations from multiple backbones are aggregated using our proposed Temporal Aggregation Module. The single-backbone Student model is then trained with bi-level fine-grained knowledge distillation framework. Right: Frame-level predictions for individual backbones vs our proposed feature aggregation method on a video of a Road Accident from the testing set of UCF-Crime.}
    \label{fig:methodcomp}
\end{figure}

Despite the extensive background of research in VAD \cite{sultani2018real, feng2021mist, li2022scale, tian2021weakly}, the development of a robust model capable of accurately detecting anomalies within videos remains a difficult task. This challenge arises from the difficulty of modeling the spatiotemporal characteristics of abnormal events, particularly those of rare occurrence and significant variability. This complexity is compounded by the labor-intensive process of collecting frame-level annotations for video data, which presents a substantial barrier towards developing an effective VAD model for real-world scenarios. Prior works on VAD have adopted a practical approach by employing weakly-supervised learning which solely requires video-level labels to develop a model capable of making frame-level predictions\cite{sultani2018real,zhang2019temporal,zhong2019graph,feng2021mist, wu2020not, zaheer2020claws, tian2021weakly, wu2021learning, li2022self, chen2022mgfn, li2022scale}. 

Although weakly-supervised VAD is an intriguing approach, it suffers from limited supervision during training, resulting from the absence of precise frame-level annotations. To overcome this challenge, previous works \cite{sultani2018real, li2022scale, tian2021weakly} have employed knowledge transfer by combining a fixed backbone, pre-trained on general video representation learning, with a dedicated prediction head to perform anomaly detection. Our exploratory evaluations, described in Figure \ref{fig:backboneabl}, highlight that knowledge transfer has a substantial impact on the performance of weakly-supervised VAD. In particular, these evaluations suggest that the impact of the knowledge transfer is even more critical than the design choice of the prediction head: {\it employing the knowledge of multiple pretrained backbones significantly enhances VAD performance.} We attribute this performance boost to the complementary nature of knowledge from different backbones resulting from variations in inductive biases and pretraining datasets. In this work, we further analyze this observation by developing an aggregated model for weakly-supervised VAD.

A careful aggregation of the knowledge from multiple models is essential especially when the training supervision is weak, \ie, video-level supervision rather than frame-level supervision. To this aim, we propose a novel Temporal Aggregation Module (TAM) that combines spatiotemporal information from the backbones through multiple self- and cross-attention mechanisms. This module comprehensively combines spatial (content) and temporal (positional) information from all the backbones to construct an effective aggregated representation of the video.    

Our empirical evaluations, discussed later in Section \ref{sec:expts}, highlight the effectiveness of this aggregated model for weakly-supervised VAD. However, this model is computationally expensive for deployment due to incorporating multiple cumbersome backbones.  To address this limitation, we develop a bi-level fine-grained knowledge distillation mechanism, which distills the knowledge from the aggregated Teacher into an efficient Student, demonstrated in Figure \ref{fig:methodcomp}. The distillation process enforces both prediction-level and feature-level similarity between the Teacher and the Student. In the former, we align the output distributions of the Teacher and Student models to capture the detailed characteristics of the aggregated predictions. In the latter case, we align the representation-level knowledge to distill more complex, higher-order feature dependencies. 

Our results highlight that the proposed TAM and knowledge distillation approach are highly beneficial for weakly supervised VAD, where the learning signal is weak.
In particular, we propose \textbf{DAKD} (\textbf{D}istilling \textbf{A}ggregated \textbf{K}nowledge with \textbf{D}isentangled Attention) that consists of a disentangled cross-attention-based Temporal Aggregation Module and a bi-level fine-grained knowledge distillation framework. 

In summary, the contributions of our work are:
\begin{enumerate}
       
        \item We argue that knowledge transfer plays an important role in the challenging setting of weakly-supervised VAD. To support this, we propose a novel Temporal Aggregation Module to effectively combine knowledge from multiple backbones.
        \item We develop a spatiotemporal knowledge distillation technique that distills the knowledge of the aggregated model into a single backbone to address the efficiency concerns. 
        \item Through extensive experiments and ablation studies, we validate the effectiveness of the proposed framework and show that our approach outperforms existing methods on benchmark datasets.
\end{enumerate} 

\begin{figure*}[t!]
    \centering
    \includegraphics[width=0.98\textwidth]{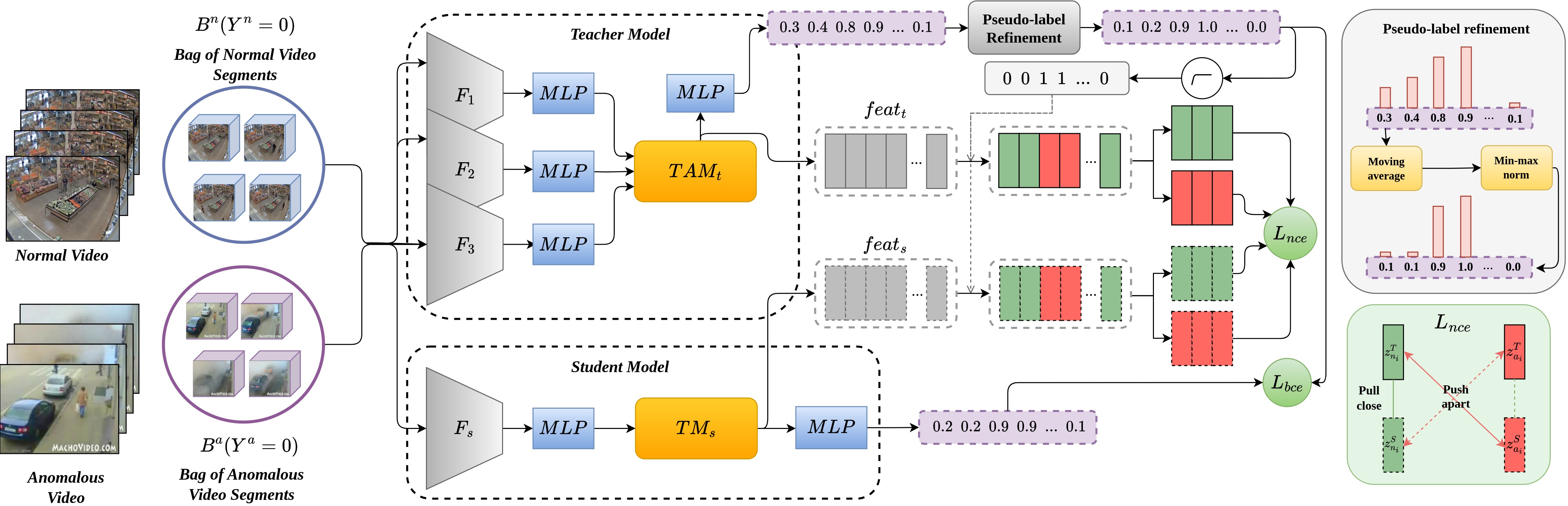}
    \caption{Schematic diagram of the proposed method. The Teacher model is initially trained with several feature extractors (Section \ref{sec:featextraction}) using the Temporal Aggregation Module (Section \ref{sec:tempnet}) in Stage 1. \textbf{Stage 2:} Feature-level and prediction-level knowledge distillation is performed to distill the knowledge of the complex Teacher model into the Student model (Section \ref{sec:distillation}).}
    \label{fig:methodology}
\end{figure*}

\section{Related Work}

Previous works on VAD can be categorized into two classes: Unsupervised VAD \cite{hasan2016learning, xu2015learning, luo2017remembering, yu2020cloze, liu2018future, lu2019future, wang2019gods, cai2021appearance, wang2020cluster, georgescu2021anomaly, liu2021hybrid, luo2017revisit, park2020learning} and Weakly-supervised VAD \cite{sultani2018real,zhang2019temporal,zhong2019graph,feng2021mist, wu2020not, zaheer2020claws, tian2021weakly, wu2021learning, li2022self, chen2022mgfn, li2022scale}. Unsupervised VAD approaches such as One-class classification assume that merely normal videos are available for training and flag videos that have a considerable deviation from the learned distribution as anomalous \cite{xu2015learning, hasan2016learning, luo2017remembering}. However, the performance of these methods is limited and often results in a high false acceptance rate. This can be attributed to the fact that normal videos with novel events closely resemble abnormal events, and it is difficult to differentiate between the two events without the related context. Weakly-supervised VAD, on the other hand, leverages video-level labels and has gained popularity for its enhanced performance.

\subsection{Weakly Supervised VAD}

Sultani \etal \cite{sultani2018real} proposed a deep Multiple-Instance Learning (MIL) framework, incorporating sparsity and temporal smoothness constraints and knowledge transfer for enhancing anomaly localization. Zhong \etal \cite{zhong2019graph} used a graph convolutional network to mitigate label noise, but had higher computational costs. Feng \etal \cite{feng2021mist} introduced a two-stage approach to fine-tune a backbone network for domain-specific knowledge. Tian \etal \cite{tian2021weakly} used top-k instances and a multi-scale temporal network for feature magnitude learning. Li \etal \cite{li2022scale} employed a scale-aware approach for capturing anomalous patterns using patch spatial relations. Zaheer \etal \cite{zaheer2020claws} minimized anomaly scores in normal regions with a Normalcy Suppression mechanism and introduced a clustering distance-based loss to improve discrimination. Despite current approaches, limited training data and weakly-supervised constraints restrict model learning. Knowledge transfer plays a crucial role in anomaly detection performance. Building upon the deep MIL framework \cite{sultani2018real}, we propose architectural changes to enhance performance on unseen data.

\subsection{Knowledge Distillation}

Knowledge distillation is a technique to transfer knowledge from a complex Teacher model to a simpler Student model. Hinton \etal \cite{hinton2015distilling} introduced the concept of aligning Teacher and Student model probabilities. FitNets\cite{romero2014fitnets} extended distillation to intermediate-level hints, focusing on matching intermediate representations. Zagoruyko and Komodakis \cite{komodakis2017paying} introduced attention-based distillation to transfer attention maps. Papernot \etal \cite{papernot2016distillation} emphasized matching intermediate representations for effective knowledge transfer. Zhang \etal \cite{zhang2020self} introduced self-distillation, leveraging the Student model as a Teacher to improve generalization. Heo \etal \cite{heo2019knowledge} improved knowledge transfer by distilling activation boundaries formed by hidden neurons.

\section{Method}

Weakly-supervised VAD aims to train models for frame-level anomaly detection using only video-level supervision. This approach faces challenges due to limited supervision and imbalanced training data, with anomalies typically occupying a small fraction of frames (e.g., 7.3\% in UCF-Crime dataset \cite{sultani2018real}). Previous works \cite{sultani2018real, zhong2019graph, tian2021weakly, zaheer2020claws, li2022scale} address this by using knowledge transfer from large video datasets. We extend this approach by utilizing multiple backbones and introducing a novel fusion method for their representations. To mitigate the increased computational demands, we propose a distillation technique to compress the aggregated model's knowledge into a single-backbone Student model. The following sections detail our feature extraction process (Section \ref{sec:featextraction}), temporal network for representation aggregation (Section \ref{sec:tempnet}), and the proposed distillation approach (Section \ref{sec:distillation}).

\begin{figure}
    \centering
\includegraphics[width=0.45\textwidth]{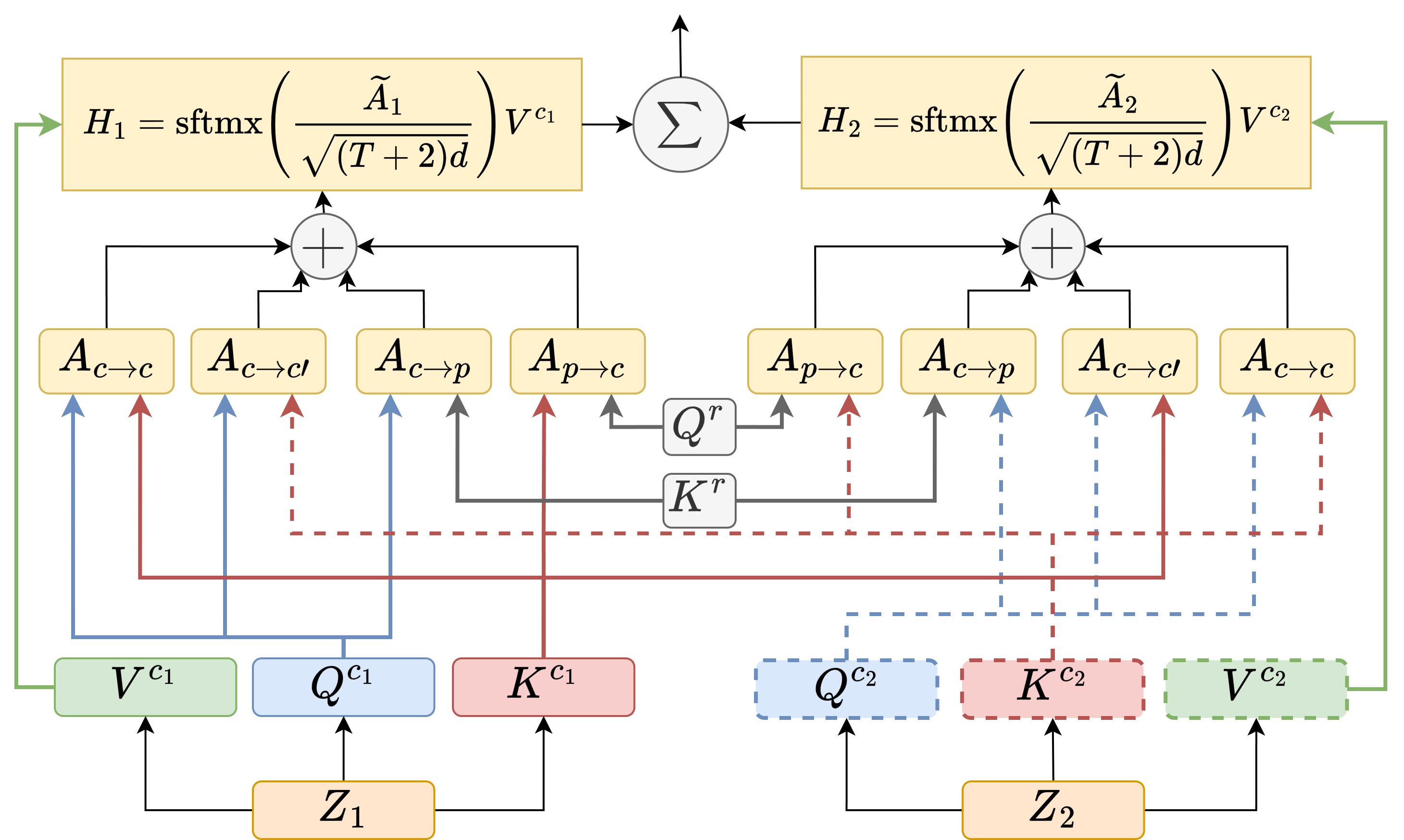}
    \caption{Schematic diagram of the proposed Temporal Aggregation Module. From the $Q^{c_t}$, $K^{c_t}$ and $V^{c_t}$ vectors obtained from the representations of the $t^{th}$ backbone and the relative position-based vectors $Q^r$ and $K^r$, four attention matrices are computed. $A_{c->c}$ is the self content-to-content attention, $A_{c->c'}$ is the cross content-to-content attention, $A_{c->p}$ is the content-to-position attention and $A_{p->c}$ is the position-to-content attention. The output value is calculated in $H_t$, and sftmx represents the softmax operation.
    }
    \label{fig:tempfig}
\end{figure}

\subsection{Feature Extraction}\label{sec:featextraction}

To alleviate the limited size of the training set and the highly imbalanced distribution of classes in weakly-supervised VAD, we adopt intensive knowledge transfer by employing multiple pre-trained video backbones for feature extraction. Different feature backbones encode different types of information, which can aid in anomaly detection and help circumvent the challenges arising from limited supervision. Additionally, different backbones can help increase the diversity of the extracted features, which can lead to a more comprehensive aggregated representation of the input videos.

Consider that the video dataset consists of $n_v$ pairs $\{(V_i, y_i)\}_{i=1}^{n_v}$, where the $i^{th}$ video, $V_i$, is a sequence of clip instances $\v_{i,j}$ and $y_i\in \{0, 1\}$ is the corresponding video-level label. Let $\psi_{1}, \dots$, $\psi_{T}$ denote the set of pre-trained backbones for extracting representations from the input videos. For each input video clip $\v_{i,j}$, we extract features using the $t^{th}$ backbone as $\z_{i,j,t}=\psi_{t}(\v_{i,j})$, where $\z_{i,j,t}\in\mathbb{R}^{d_t}$ and $d_t$ denotes the cardinality of the output of the $t^{th}$ backbone. After extracting representations using multiple backbones, we aggregate them using a novel Temporal Aggregation Module described in the next section.

\subsection{Temporal Aggregation Module}
\label{sec:tempnet}

In weakly-supervised VAD, incorporating relative positional information is vital due to the low-pass temporal frequency characteristics of natural events, where anomalous frames cluster together rather than appearing sporadically. Leveraging this information enhances anomaly detection performance, which we achieve by utilizing a disentangled attention mechanism that inherently accounts for relative positional information during attention computation. This mechanism \cite{he2020deberta} employs a relative positional bias, with the maximum relative distance parameterized by $k$. The relative distance between positions $i$ and $j$ is encoded by the function $\gamma(i,j)$, constrained within $[0,2k)$, thereby reducing attention model complexity and making it suitable for low-data and weak-supervision scenarios. Formally, $\gamma(i,j)$ is defined as follows:

\begin{equation}
\gamma(i,j) =
\begin{cases}
0 & \text{for $i - j \leq -k $ },\\
2k - 1 & \text{for $i - j \geq k $},\\
i - j + k & \text{others}.
\end{cases}
\label{eqn: rel_pos}
\end{equation}
The disentangled attention has three attention components: i) {\it content-to-content attention}: This component attends to the content of a token at position {\it i} by interacting with the content of the token at position {\it j} within the same sequence, ii) {\it content-to-position attention}: This attention component considers the content of token {\it i} and its relative position to token {\it j}, capturing how the content of token {\it i} influences its attention weight concerning token {\it j}. iii) {\it position-to-content attention}: Similarly, this component assesses the content of token {\it j} and its relative position to token {\it i}, elucidating how the content at position {\it j} influences its attention weight with respect to token {\it i}. 

We further disentangle this attention mechanism to adopt it for multi-input scenarios. To this aim, we add a cross-attention module to fuse information from multiple input sequences. Table \ref{tab:attention} evaluates the contribution of each of these attention components. Given $T$ input sequences $Z_t$, $t\in \{1, \dots, T\}$, we define the query, key, and value for each of the representations as:    
\begin{equation}
\begin{split}
Q^{c_t}=Z_tW_{q, c_t}, \quad K^{c_t}=Z_tW_{k, c_t}, \quad V^{c_t}=Z_tW_{v, c_t}.
\end{split}
\label{eq:6}
\end{equation}
The shared relative position key and query are also computed as :
\begin{equation}
    Q^{r} = \tilde{Z}W_{q,r}, \quad K^{r} = \tilde{Z}W_{k,r},
\end{equation}
where $\tilde{Z} \in \mathbb{R}^{2k \times d}$ represents the relative position embedding vectors shared across all layers and backbones (\ie, staying fixed during forward propagation).

Our aggregation attention mechanism is then formulated as:
\begin{equation}
\begin{aligned}
    \tilde{A}_{i,j,t} = & \underbrace{ Q_i^{c_t}{K_j^{c_t}}^\top}_{\text{(self content-to-content)}} + \underbrace{ \sum_{h,h\neq t} Q_i^{c_t}{K_j^{c_h} }^\top}_{\text{(cross content-to-content)}} \\
    & + \underbrace{Q_i^{c_t}{K^r_{\gamma(i,j)}}^\top}_{\text{(content-to-position)}} + \underbrace{ K_j^{c_t}{Q^r_{\gamma(j,i)}}^\top}_{\text{(position-to-content)}},
\end{aligned}
\label{eq:3}
\end{equation}
and the output is computed as: 
\begin{equation}
    H_t = softmax\big(\dfrac{\tilde{A}_t}{\sqrt{(T+2)d}} \big) V^{c_t},
\end{equation}
where $T$ is the number of backbones, and the final aggregated output is $H=\tfrac{1}{T}\sum_t H_t$.
$Q^{c_t}$, $K^{c_t}$, and $V^{c_t}$ are content vectors derived through projection matrices $W_{q, c_t}$, $W_{k, c_t}$, $W_{v, c_t}$ $\in \mathbb{R}^{ d \times d}$ and $t \in \{1, \dots, T\}$ is the index of the backbone. $Q_{r}$ and $K_{r}$ correspond to the projected relative position vectors, facilitated by projection matrices $W_{q,r}$ and $W_{k,r} \in \mathbb{R}^{d \times d}$, respectively. The architecture of the Temporal Aggregation Module integrates the aforementioned disentangled attention mechanism to improve the relative encoding and fusing of representations from multiple backbones.

\subsection{Bi-level Fine-grained Knowledge Distillation}\label{sec:distillation}
 
Our approach leverages the knowledge from multiple pre-trained backbones, allowing it to benefit from their collective expertise. We employ the MIL ranking approach, proposed by Sultani \etal \cite{sultani2018real} for training the first stage aggregated model. This intensive knowledge transfer aims to address the scarcity of supervision, which often hinders learning in the current weakly-supervised learning setup. However, using multiple backbones drastically increases computational overhead, thereby making the model less suitable for real-world applications. To overcome these issues, we develop a knowledge distillation approach, as shown in Figure \ref{fig:methodology}, that distills the knowledge of the aggregated Teacher at prediction and representation levels into the Student model.\\ 

\noindent\textbf{Prediction-level  distillation:}
In the first level of distillation, we align the output distributions of the Teacher and Student models using the cross-entropy loss function \cite{hinton2015distilling}. Weakly-supervised VAD methods \cite {sultani2018real, li2022scale, tian2021weakly} generally use a single segment or top-k segments for the given input during the training since they solely have access to the video-level annotation. However, for distillation, despite the lack of fine-grained ground truth labels, we use the Teacher's segment-level predictions to provide robust learning signals. These predictions act as soft pseudo-labels for training the Student model. The trained aggregated model generates scores for anomalous videos marked as $S^a = \{s_{i}^{a}\}_{i = 1}^{n_s}$, where $n_s$ is the number of segments. Based upon \cite{feng2021mist}, to remove the jitter and refine the anomaly scores, we use a convolutional kernel of size $\epsilon$ as a moving average filter and use min-max normalization afterward. Min-max normalization helps to focus on the anomalous segments during training. The impact of using moving average filter and min-max normalization is presented in Table \ref{tab:table_comp}. Min-max normalization and moving average filter are described as:

\begin{equation}
\label{eq: pseudoref}
\begin{split}
\hat{y}_{i}^{a} &= \frac{\tilde{s}_{i}^{a} - \min (\tilde{S}^{a})}{\max (\tilde{S}^{a}) - \min (\tilde{S}^{a})},\quad i \in [1, n_s], \\
\tilde{s}_{i}^{a} &= \frac{1}{2\epsilon} \sum_{j = i - \epsilon}^{i + \epsilon} {s}_{j}^{a},
\end{split}
\end{equation}
\noindent respectively, where $\min$ and $\max$ functions compute the minimum and maximum scores in the given set.

We refine the anomaly scores into $Y^a = \{\hat{y}_{i}^{a}\}_{i = 1}^{n_s}$ and use these as soft pseudo-labels. Since we are certain about the segment-level annotation in the normal videos, we can combine the soft anomaly labels with normal videos. Given the nature of the VAD task, which typically involves predictions for two classes, the conventional posterior matching approach encounters limitations due to the limited support of the distribution. To mitigate this issue, we extend the distillation process to operate at the feature level, employing the InfoNCE loss \cite{oord2018representation, tian2019contrastive}. \\

\noindent\textbf{Feature-level  distillation:}
In the context of feature-level distillation, we utilize a multilayer perceptron (MLP) with a single hidden layer to transform the input representations $\h_i$ from both the Student and Teacher models into corresponding feature vectors $\z_i = g(\h_i) = W^{(2)}\sigma(W^{(1)}\h_i)$, with $\sigma$ representing the ReLU nonlinearity\cite{agarap2019relu}. Subsequently, leveraging the Teacher model's prediction outputs, we determine class labels for individual features by applying a threshold $\delta$. This facilitates the identification of four distinct feature subsets: $\z_a^T$ (anomaly features of the Teacher), $\z_n^T$ (normal features of the Teacher), $\z_a^S$ (anomaly features for the Student), and $\z_n^S$ (normal features of the Student). Here, the superscripts {\it T} and {\it S} denote the Teacher and Student models, respectively. We utilize cosine similarity, denoted as $\text{sim}(.,.)$, as a measure of similarity between input vectors. 

Our feature distillation loss using InfoNCE is defined as:
\begin{equation}
\begin{split}
    \mathcal{L}_{nce} = -\log \dfrac{e^{\text{sim}(\z_{a_i}^T, \z_{a_i}^S)/\tau}}{\sum_{k=1}^{N}e^{\text{sim}(\z_{a_i}^T, \z_{n_k}^S)/\tau} + e^{\text{sim}(\z_{a_i}^T, \z_{a_i}^S)/\tau}} \\
     -\log \dfrac{e^{\text{sim}(\z_{n_i}^T, \z_{n_i}^S)/\tau}}{\sum_{k=1}^{N}e^{\text{sim}(\z_{n_i}^T, \z_{a_k}^S)/\tau} + e^{\text{sim}(\z_{n_i}^T, \z_{n_i}^S)/\tau}},
\end{split}
\end{equation}
where $\tau$ represents the temperature parameter.
The complete loss function for the distillation is as follows:
\begin{equation}
\mathcal{L}_d = \mathcal{L}_{bce}(y^T, y^S) + \alpha\tau^2 \mathcal{L}_{nce}(T, S),
\label{eq: losseqn}
\end{equation}
where $\mathcal{L}_{nce}$ represents the feature-level distillation loss using InfoNCE, $\mathcal{L}_{bce}$ represents the BCE loss, $\mathcal{L}_d$ represents the combined loss for distillation, and $\alpha$ represents the scaling coefficient to control the contribution of the two loss terms.

\section{Experiments}
\label{sec:expts}

\subsection{Datasets and Metrics}

Our model is evaluated on three benchmark datasets for weakly-supervised VAD: UCF-Crime, ShanghaiTech, and XD-Violence. The UCF-Crime dataset \cite{sultani2018real} contains 1900 untrimmed videos totaling 128 hours, captured by surveillance cameras in diverse real-world settings, with 13 types of anomalies. The ShanghaiTech dataset \cite{liu2018future} comprises 437 videos from fixed-angle street cameras, featuring 13 background scenes. We follow Zhong \etal's \cite{zhong2019graph} approach to adapt it for weakly-supervised learning. The XD-Violence dataset \cite{wu2020not} is a comprehensive multiscene collection sourced from various media, containing 4754 untrimmed videos spanning over 217 hours. All three datasets provide video-level labels for training and frame-level labels for testing, allowing for robust evaluation of weakly-supervised VAD models across diverse scenarios and anomaly types.

\begin{table}
\small
\centering
\begin{tabular}{lc}
\textbf{Temporal Network} & \textbf{AUC(\%)} \\
\hline
MTN & 85.31 \\
Multihead Attention & 86.28 \\
LSTM & 86.97 \\ 
RNN & 86.99 \\
Disentangled Attention & 87.09 \\
GRU & 87.53 \\
1D CNN & 87.72 \\
% $\text{TAM}_{RNN}$ & 88.18 \\
% $\text{TAM}_{GRU}$ & 88.24 \\ 
\textbf{TAM} & \textbf{88.34} \\ \hline
\end{tabular}
\caption{Comparison of the proposed Temporal Aggregation Module ($\text{TAM}$) with other variants including the MTN module from RTFM. We observe that TAM is superior to other temporal models at capturing spatio-temporal dependencies.}
\label{tab:tempnet}
\end{table}

\noindent\textbf{Evaluation Metrics:} In order to assess the effectiveness of our approach, we utilize the frame-based receiver operating characteristic (ROC) curve and the area under the curve (AUC), which have been commonly used in previous studies on anomaly detection \cite{sultani2018real, tian2021weakly, feng2021mist}. Based on \cite{wu2020not}, we use average precision (AP) as the evaluation measure for the XD-Violence dataset.

\begin{table}[t]
\small
\centering
\begin{tabular}{llcc}
\textbf{Method}               & \textbf{Features}      & \textbf{T=32} & \textbf{AUC (\%)}  \\  \hline
Sultani \etal  \cite{sultani2018real}                 & C3D-RGB       & \checkmark & 75.41                           \\
Sultani \etal  \cite{sultani2018real}                   & I3D-RGB       & \checkmark & 77.92                           \\
Zhang \etal  \cite{zhang2019temporal}                  & C3D-RGB       & \checkmark & 78.66                           \\
% Zaheer et al. (2022) \cite{zaheer2020claws}                 & ResNext       & -                         & 79.84                           \\
GCN  \cite{zhong2019graph}                           & TSN-RGB       & -                         & 82.12                           \\
MIST  \cite{feng2021mist}                          & I3D-RGB       & -                         & 82.30                           \\
Wu \etal     \cite{wu2020not}                 & I3D-RGB       & \checkmark & 82.44                           \\
CLAWS     \cite{zaheer2020claws}                     & C3D - RGB     & -                         & 83.03                           \\
RTFM$^*$  \cite{tian2021weakly}                       & I3D-RGB       & \checkmark & 84.30                           \\
Wu \etal  \cite{wu2021learning}                    & I3D-RGB       & \checkmark & 84.89                           \\
MSL    \cite{li2022self}                         & I3D-RGB       & \checkmark & 85.30                           \\
MSL    \cite{li2022self}                         & VSwin-RGB & \checkmark & 85.62                           \\
S3R      \cite{wu2022self}                      & I3D-RGB       & \checkmark & 85.99                           \\
SSRL$^*$   \cite{li2022scale}                      & I3D-RGB       & \checkmark & 86.79                           \\
MGFN  \cite{chen2022mgfn}                          & I3D-RGB       & \checkmark & 86.98                           \\
\textbf{DAKD$_{T}$ (Ours)} & \textbf{Multiple}      & \textbf{\checkmark} & \textbf{88.15} \\
\textbf{DAKD$_{S}$ (Ours)} & \textbf{I3D}      & \textbf{\checkmark} & \textbf{88.10} \\
\textbf{DAKD$_{S}$ (Ours)} & \textbf{CLIP}      & \textbf{\checkmark} & \textbf{88.34} \\
\hline
\end{tabular}
\caption{Comparison with existing weakly-supervised methods on UCF-Crime dataset. DAKD$_{T}$ and DAKD$_{S}$ denote Teacher and Student models. T=32 indicates 32 non-overlapping video segments. Features column shows backbone used for feature extraction. Asterisk (*) indicates methods for which we could not validate the performance using the official code or our implementation. Results are the average over five independent runs. }
\label{tab:ucf_result}
\end{table}

\subsection{Implementation Details}
Our proposed method is implemented using PyTorch \cite{paszke2019pytorch}. We divide each video into 32 non-overlapping segments to pass through the feature extractors.\\

\noindent\textbf{Teacher Model:} For the Teacher Model, we utilize the I3D \cite{carreira2017quo}, S3D \cite{zhang2018s3d}, and CLIP \cite{radford2021clip} backbones to obtain representations for the video inputs. Before aggregation, the feature vectors are projected to a common dimension (512) using two-layer MLPs with ReLU \cite{agarap2019relu} activation in the first layer. The input dimension for the MLPs processing I3D and S3D features is 1024, while for the one processing CLIP representations is 512. The hidden dimension is 512 for each of the three MLPs. 

In our Temporal Aggregation Module, we utilize disentangled attention along with our proposed cross-attention mechanism to combine multiple inputs, as explained in Section \ref{sec:tempnet}. The disentangled attention module has one hidden layer with a hidden embedding dimension of 1024 and 8 attention heads. Notably, the TAM shares positional embeddings among all backbones. We experimented with values of maximum relative distance $k$ from 1 to 32, where 32 is the maximum segment index, to determine the optimal value that constrains the maximum distance between two positions $(i,j)$ in disentangled attention. The aggregated features are then passed through a feedforward network with hidden dimensions of 512 and 32, and with a sigmoid activation in the final layer to obtain the segment-level prediction. The conventional MIL Loss \cite{sultani2018real} serves as the loss function during training.\\

\noindent\textbf{Student Model:} In the Student Model, we use the CLIP \cite{radford2021clip} backbone which has shown notable performance in video analysis tasks \cite{lin2022frozen}. We then pass the representations to the Temporal Module. The Temporal Module uses the disentangled attention mechanism with the omission of cross content-to-content attention term. The attention normalization factor is also adjusted based on the single backbone formulation. The dimensionality of embeddings and the number of attention heads is similar to Teacher model formulation. We train the Student model with the bi-level fine-grained knowledge distillation approach as discussed in Section \ref{sec:distillation}. Employing contrastive loss for representation-level distillation, we mask the positive and negative examples using the threshold $\delta = 0.9$ on the Teacher's predictions. Then we calculate the cosine similarity between the samples from the Teacher and Student and obtain the loss for cross-positive and cross-negative examples. For prediction-level distillation, we use the BCE loss function to align the output distributions. Finally, we use a linear combination of the two loss terms using the parameter $\alpha$ to calculate the final loss. We also scale the $\mathcal{L}_{nce}$ loss by $\tau^2$ (Equation \ref{eq: losseqn}) which results in better training.

The Teacher and Student models are trained for 100 epochs, using the Adagrad optimizer \cite{duchi2011adaptive} with a weight decay of 0.001 and a learning rate of 0.0001 for the temporal models and 0.001 otherwise. The training batch size is 60, and each batch consists of 30 normal and 30 anomalous video clips.

\begin{figure}
    \centering
    \includegraphics[width=0.4\textwidth]{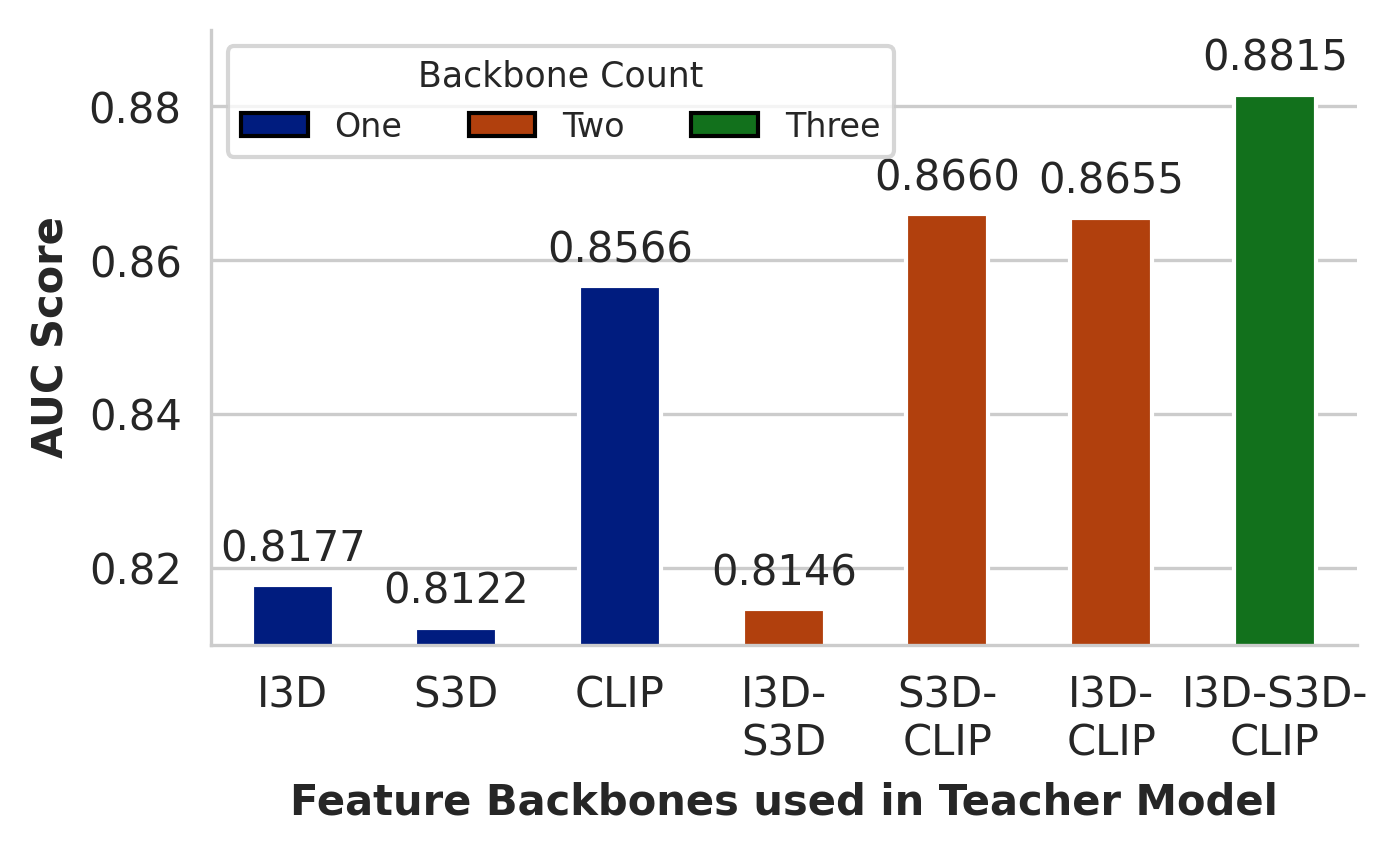}
    \captionof{figure}{Ablation study on the UCF-Crime dataset to investigate the impact of feature backbones used in the Teacher Model. We observe that the involvement of the CLIP backbone significantly boosts the AUC score. The combination of jointly using all three backbones (I3D, S3D, and CLIP) provides the best performance.}
    \label{fig:backboneabl}
\end{figure}

\subsection{Comparison with the state of the art}
Table \ref{tab:ucf_result} presents our main results on the UCF-Crime dataset, while Table  \ref{tab:sht_result} shows the results for the ShanghaiTech and XD-Violence datasets.
DAKD$_{T}$ and DAKD$_{S}$ outperform all the existing weakly-supervised methods by a significant margin on all the datasets. Remarkably, on the UCF-Crime dataset, DAKD$_{S}$ outperforms current SOTA methods, MGFN \cite{chen2022mgfn} by 1.36\%, SSRL \cite{li2022scale} by 1.55\%, S3R \cite{wu2022self} by 2.35\%, and RTFM \cite{tian2021weakly} by 4.04\%. DAKD$_{S}$ is also extremely efficient compared to SSRL, which uses multi-scale video crops for training, making it suitable for real-world use. Additionally, DAKD also achieves an AUC score of 98.10\% on the ShanghaiTech dataset, providing superior performance although the performance of previous methods seems to be saturated on this dataset. Moreover, DAKD achieves an AP score of 85.61\% on the XD-Violence dataset and outperforms existing methods by a significant margin.

\begin{figure*}[t!]
    \centering
    \includegraphics[width=0.98\textwidth]{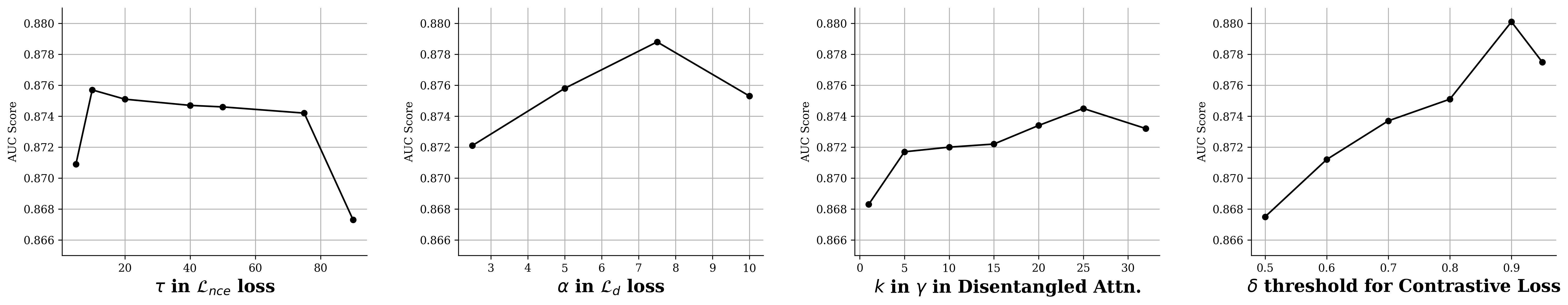}
    \caption{Ablation studies performed on major hyperparameters including the temperature for the contrastive loss $\tau$, the coefficient of the total distillation loss $\alpha$, the maximum relative distance parameter $k$ in the disentangled attention mechanism, and the threshold $\delta$ used to determine class labels for the contrastive loss. The ablations are performed on the UCF-Crime dataset.}
    \label{fig:studentabl}
\end{figure*}

\begin{figure}
    \centering
    \includegraphics[width=0.45\textwidth]
    {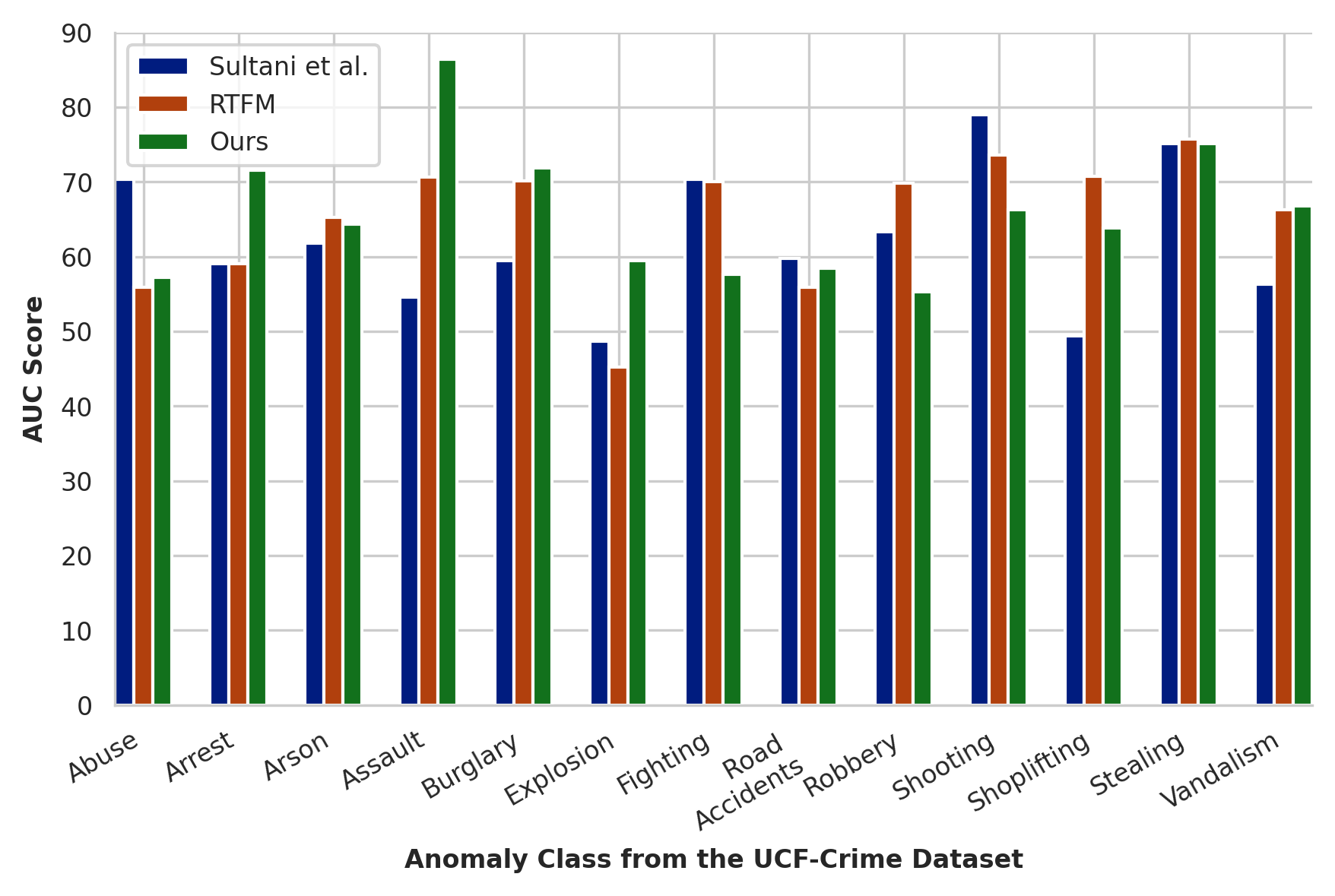}
    \captionof{figure}{AUC Scores with respect to individual anomaly classes on the UCF-Crime dataset. We compare our results with Sultani \etal \cite{sultani2018real} and RTFM \cite{tian2021weakly} and observe significant improvements in multiple classes, notably Assault, Arrest, Burglary and Explosion.}
    \label{fig:multiclass}
\end{figure}

We further analyze the performance of the models on each anomaly class in UCF-Crime to highlight the effectiveness of DAKD. 
Figure \ref{fig:multiclass} presents the class-wise AUC scores for the Anomaly classes in UCF-Crime of our method compared with that of Sultani \etal \cite{sultani2018real} and RTFM \cite{tian2021weakly}. Our approach outperforms existing methods by a significant margin in classes such as Assault, Arrest, Burglary, Explosion, and Vandalism. 

\subsection{Ablation Study}
\label{sec:ablation}

\noindent\textbf{Analysis of Different Feature Backbones:} Ablation studies were conducted to assess the impact of different pre-trained feature backbones on our Teacher Model's training. Specifically, we utilized three feature backbones: I3D \cite{carreira2017quo}, S3D \cite{zhang2018s3d}, and CLIP \cite{radford2021clip}. The results of these ablation studies can be found in Figure \ref{fig:backboneabl}. It is evident from the figure that configurations involving the CLIP backbone consistently outperform other combinations. Notably, the combination involving all three backbones yields the best performance and is consequently adopted for training the Teacher Model. This choice not only enhances the diversity of input features but also mitigates the challenges posed by the limited availability of training data.\\

\begin{table}
\small
\centering
\begin{tabular}{llccc}
\textbf{Method}               & \textbf{Features}      & \textbf{T=32} & \textbf{SHT} & \textbf{XDV} \\ 
\hline
 Sultani \etal \cite{sultani2018real} & C3D-RGB       & \checkmark & 86.30 & 73.20 \\
Zhang \etal  \cite{zhang2019temporal}    & C3D-RGB       & \checkmark & 82.50 & - \\
MIST  \cite{feng2021mist}        & I3D-RGB       & -    & 94.83  &  - \\
CLAWS  \cite{zaheer2020claws}        & C3D - RGB     & -    & 89.67 & -\\
RTFM$^*$  \cite{tian2021weakly}         & I3D-RGB       & \checkmark & 97.21 & 77.81\\
Wu \etal \cite{wu2020not}    & I3D-RGB       & \checkmark & - & 78.64 \\
Wu \etal \cite{wu2021learning}    & I3D-RGB       & \checkmark & 97.48 & -\\
MSL    \cite{li2022self}           & I3D-RGB       & \checkmark & 96.08 & 78.28\\
SSRL$^*$ \cite{li2022scale}          & I3D-RGB       & \checkmark & 97.04 & -\\
MSL    \cite{li2022self}           & VSwin-RGB & \checkmark & 97.32 & 78.59\\
\textbf{DAKD$_{T}$ (Ours)} & \textbf{Multiple}      & \textbf{\checkmark} & \textbf{98.08} & \textbf{84.78}\\
\textbf{DAKD$_{S}$ (Ours)} & \textbf{I3D}      & \textbf{\checkmark} & \textbf{98.02} & \textbf{85.12}\\
\textbf{DAKD$_{S}$ (Ours)} & \textbf{CLIP}      & \textbf{\checkmark} & \textbf{98.10} & \textbf{85.61}\\
\hline
\end{tabular}
\caption{Performance comparison with existing weakly-supervised methods on ShanghaiTech (SHT, AUC score) and XD-Violence (XDV, AP score) datasets. DAKD$_{T}$ and DAKD$_{S}$ denote Teacher and Student models. Other notations as in Table \ref{tab:ucf_result}.}
\label{tab:sht_result}
\end{table}

\noindent\textbf{Analysis of Different Temporal Modules:} We study the impact of the proposed Temporal Aggregation Module compared to prominent temporal networks. The feature representations obtained from the  I3D \cite{carreira2017quo}, S3D \cite{zhang2018s3d}, and CLIP \cite{radford2021clip} backbones are passed to the specified temporal module. The particular results are presented in Table \ref{tab:tempnet}. From Table \ref{tab:tempnet}, we can see that the proposed TAM specification outperforms other popular temporal networks in terms of the AUC score. Notably, TAM outperforms vanilla disentangled attention and multihead attention mechanisms showing its efficacy.\\

\noindent\textbf{Analysis of Other Parameters:} We investigated the effects of several key training parameters, as shown in Figure \ref{fig:studentabl} and Table \ref{tab:table_comp}. Increasing the temperature $\tau$ in the $\mathcal{L}_{nce}$ loss beyond $\tau = 10$ reduced AUC scores, as higher $\tau$ weakens penalties on hard negatives, while smaller $\tau$ enhances feature separation. The scaling factor $\alpha$, which balances $\mathcal{L}_{nce}$ in the overall loss $\mathcal{L}_d$, achieved the highest AUC at $\alpha = 7.5$, with higher values prioritizing feature-level distillation objective. Ablation studies on the maximum relative distance $k$ in $\gamma(i,j)$ showed optimal performance at $k = 25$, balancing attention model complexity and generalization to higher number of segments. Similarly, $\delta = 0.9$ yielded the best results for assigning class-based labels to the segment-level features in feature-level distillation objective. In Table \ref{tab:table_comp}, we also compare the impact of various key components of our framework like TAM, the bi-level distillation objective, and the pseudo-label refinement. The results highlight the importance of each component and show that all the mentioned components are crucial towards achieving optimal performance.

\begin{table}[h!]
\centering
\small
\begin{tabular}{llc}
\textbf{Method} & \textbf{UCF} & \textbf{SHT} \\ \hline
Baseline & 77.92   & 86.30                                    \\
Ours w/o TAM  & 82.27 & 91.34                                    \\
Ours w/o $\mathcal{L}_{nce}$      & 86.91                                    & 96.52                                    \\
Ours w/o $\mathcal{L}_{bce}$      & 87.60                                    & 96.89                                    \\ 
Ours w/o Min-Max Normalization     & 87.90                                    & 97.21                                   \\ 
Ours w/o Moving Average Filter     & 88.10                                    & 97.63                                    \\ 
\textbf{Ours}    & \textbf{88.34}          & \textbf{98.10} \\ \hline 
\end{tabular}
\caption{Ablation studies on UCF-Crime and ShanghaiTech datasets, examining key components of our framework. Baseline: MIL method \cite{sultani2018real}. Variants: without TAM, representation-level distillation ($\mathcal{L}_{nce}$), and prediction-level distillation ($\mathcal{L}_{bce}$). Also includes ablations on min-max normalization and moving average filter for pseudo-label refinement.}
\label{tab:table_comp}
\end{table}

\begin{table}[t!]
\centering
\small
\begin{tabular}{lcc}
\textbf{Attention Mechanism} & \textbf{UCF} & \textbf{SHT} \\ \hline
Content-to-Position      & 86.12                                    & 95.72                                    \\
Position-to-Content      & 86.47                                    & 96.34                                    \\ 
Cross Content-to-Content                     & 86.81                                    & 96.58                                    \\
Self Content-to-Content                          & 87.56                                    & 97.46                                    \\
\textbf{All Components}    & \textbf{88.34}          & \textbf{98.10} \\ \hline       
\end{tabular}
\caption{Ablation studies on the components of the aggregation attention mechanism described in Section \ref{sec:tempnet}. The table presents the AUC scores for the UCF-Crime (UCF) and ShanghaiTech (SHT) datasets.}
\label{tab:attention}
\end{table}

\subsection{Qualitative Results}
From Figure \ref{fig:methodcomp}, it is clear that while individual backbones struggle at corresponding to the ground truth frame-level annotations, our approach of using aggregated features is able to correctly localize the anomaly. The combined features leverage the power of individual backbones and show effective performance where dataset size is limited. Our approach also shows a comparatively smoother transition between normal and anomalous regions, demonstrating that it is able to consider the temporal localization of an event.

\section{Conclusion}

In this work, we introduced DAKD to address weakly-supervised VAD challenges, particularly the scarcity of frame-level labeled data. Our approach features a Temporal Aggregation Module (TAM) that combines diverse representations from multiple backbones using disentangled cross-attention. To mitigate computational costs, DAKD employs a bi-level knowledge distillation mechanism, transferring the aggregated model's knowledge to a single-backbone Student. Extensive evaluations on UCF-Crime, ShanghaiTech, and XD-Violence datasets demonstrate the effectiveness of our aggregated model and show that the distilled Student consistently outperforms existing methods, achieving state-of-the-art performance in weakly-supervised VAD.

\section{Acknowledgement}

This work was supported in part by U.S. NIH grants R01GM134020 and P41GM103712, NSF grants DBI-1949629, DBI-2238093, IIS-2007595, IIS-2211597, and MCB-2205148. This work was supported in part by Oracle Cloud credits and related resources provided by Oracle for Research, and the computational resources support from AMD HPC Fund.

{
    \small
    \bibliographystyle{ieee_fullname}
    \bibliography{egbib}

\begin{thebibliography}{10}\itemsep=-1pt

\bibitem{agarap2019relu}
Abien~Fred Agarap.
\newblock Deep learning using rectified linear units (relu), 2019.

\bibitem{cai2021appearance}
Ruichu Cai, Hao Zhang, Wen Liu, Shenghua Gao, and Zhifeng Hao.
\newblock Appearance-motion memory consistency network for video anomaly detection.
\newblock In {\em Proceedings of the AAAI conference on artificial intelligence}, volume~35, pages 938--946, 2021.

\bibitem{carreira2017quo}
Joao Carreira and Andrew Zisserman.
\newblock Quo vadis, action recognition? a new model and the kinetics dataset.
\newblock In {\em proceedings of the IEEE Conference on Computer Vision and Pattern Recognition}, pages 6299--6308, 2017.

\bibitem{chen2022mgfn}
Yingxian Chen, Zhengzhe Liu, Baoheng Zhang, Wilton Fok, Xiaojuan Qi, and Yik-Chung Wu.
\newblock Mgfn: Magnitude-contrastive glance-and-focus network for weakly-supervised video anomaly detection.
\newblock {\em arXiv preprint arXiv:2211.15098}, 2022.

\bibitem{duchi2011adaptive}
John Duchi, Elad Hazan, and Yoram Singer.
\newblock Adaptive subgradient methods for online learning and stochastic optimization.
\newblock {\em Journal of machine learning research}, 12(7), 2011.

\bibitem{feng2021mist}
Jia-Chang Feng, Fa-Ting Hong, and Wei-Shi Zheng.
\newblock Mist: Multiple instance self-training framework for video anomaly detection.
\newblock In {\em Proceedings of the IEEE/CVF conference on computer vision and pattern recognition}, pages 14009--14018, 2021.

\bibitem{georgescu2021anomaly}
Mariana-Iuliana Georgescu, Antonio Barbalau, Radu~Tudor Ionescu, Fahad~Shahbaz Khan, Marius Popescu, and Mubarak Shah.
\newblock Anomaly detection in video via self-supervised and multi-task learning.
\newblock In {\em Proceedings of the IEEE/CVF conference on computer vision and pattern recognition}, pages 12742--12752, 2021.

\bibitem{hasan2016learning}
Mahmudul Hasan, Jonghyun Choi, Jan Neumann, Amit~K Roy-Chowdhury, and Larry~S Davis.
\newblock Learning temporal regularity in video sequences.
\newblock In {\em Proceedings of the IEEE conference on computer vision and pattern recognition}, pages 733--742, 2016.

\bibitem{he2020deberta}
Pengcheng He, Xiaodong Liu, Jianfeng Gao, and Weizhu Chen.
\newblock Deberta: Decoding-enhanced bert with disentangled attention.
\newblock {\em arXiv preprint arXiv:2006.03654}, 2020.

\bibitem{heo2019knowledge}
Byeongho Heo, Minsik Lee, Sangdoo Yun, and Jin~Young Choi.
\newblock Knowledge transfer via distillation of activation boundaries formed by hidden neurons.
\newblock In {\em Proceedings of the AAAI Conference on Artificial Intelligence}, volume~33, pages 3779--3787, 2019.

\bibitem{hinton2015distilling}
Geoffrey Hinton, Oriol Vinyals, and Jeff Dean.
\newblock Distilling the knowledge in a neural network.
\newblock {\em arXiv preprint arXiv:1503.02531}, 2015.

\bibitem{komodakis2017paying}
Nikos Komodakis and Sergey Zagoruyko.
\newblock Paying more attention to attention: improving the performance of convolutional neural networks via attention transfer.
\newblock In {\em ICLR}, 2017.

\bibitem{li2022scale}
Guoqiu Li, Guanxiong Cai, Xingyu Zeng, and Rui Zhao.
\newblock Scale-aware spatio-temporal relation learning for video anomaly detection.
\newblock In {\em Computer Vision--ECCV 2022: 17th European Conference, Tel Aviv, Israel, October 23--27, 2022, Proceedings, Part IV}, pages 333--350. Springer, 2022.

\bibitem{li2022self}
Shuo Li, Fang Liu, and Licheng Jiao.
\newblock Self-training multi-sequence learning with transformer for weakly supervised video anomaly detection.
\newblock In {\em Proceedings of the AAAI Conference on Artificial Intelligence}, volume~36, pages 1395--1403, 2022.

\bibitem{lin2022frozen}
Ziyi Lin, Shijie Geng, Renrui Zhang, Peng Gao, Gerard de Melo, Xiaogang Wang, Jifeng Dai, Yu Qiao, and Hongsheng Li.
\newblock Frozen clip models are efficient video learners.
\newblock In {\em European Conference on Computer Vision}, pages 388--404. Springer, 2022.

\bibitem{liu2018future}
Wen Liu, Weixin Luo, Dongze Lian, and Shenghua Gao.
\newblock Future frame prediction for anomaly detection--a new baseline.
\newblock In {\em Proceedings of the IEEE conference on computer vision and pattern recognition}, pages 6536--6545, 2018.

\bibitem{liu2021hybrid}
Zhian Liu, Yongwei Nie, Chengjiang Long, Qing Zhang, and Guiqing Li.
\newblock A hybrid video anomaly detection framework via memory-augmented flow reconstruction and flow-guided frame prediction.
\newblock In {\em Proceedings of the IEEE/CVF International Conference on Computer Vision}, pages 13588--13597, 2021.

\bibitem{lu2019future}
Yiwei Lu, K~Mahesh Kumar, Seyed shahabeddin Nabavi, and Yang Wang.
\newblock Future frame prediction using convolutional vrnn for anomaly detection.
\newblock In {\em 2019 16th IEEE International Conference on Advanced Video and Signal Based Surveillance (AVSS)}, pages 1--8. IEEE, 2019.

\bibitem{luo2017remembering}
Weixin Luo, Wen Liu, and Shenghua Gao.
\newblock Remembering history with convolutional lstm for anomaly detection.
\newblock In {\em 2017 IEEE International Conference on Multimedia and Expo (ICME)}, pages 439--444. IEEE, 2017.

\bibitem{luo2017revisit}
Weixin Luo, Wen Liu, and Shenghua Gao.
\newblock A revisit of sparse coding based anomaly detection in stacked rnn framework.
\newblock In {\em Proceedings of the IEEE international conference on computer vision}, pages 341--349, 2017.

\bibitem{oord2018representation}
Aaron van~den Oord, Yazhe Li, and Oriol Vinyals.
\newblock Representation learning with contrastive predictive coding.
\newblock {\em arXiv preprint arXiv:1807.03748}, 2018.

\bibitem{papernot2016distillation}
Nicolas Papernot, Patrick McDaniel, Xi Wu, Somesh Jha, and Ananthram Swami.
\newblock Distillation as a defense to adversarial perturbations against deep neural networks.
\newblock In {\em 2016 IEEE symposium on security and privacy (SP)}, pages 582--597. IEEE, 2016.

\bibitem{park2020learning}
Hyunjong Park, Jongyoun Noh, and Bumsub Ham.
\newblock Learning memory-guided normality for anomaly detection.
\newblock In {\em Proceedings of the IEEE/CVF conference on computer vision and pattern recognition}, pages 14372--14381, 2020.

\bibitem{paszke2019pytorch}
Adam Paszke, Sam Gross, Francisco Massa, Adam Lerer, James Bradbury, Gregory Chanan, Trevor Killeen, Zeming Lin, Natalia Gimelshein, Luca Antiga, et~al.
\newblock Pytorch: An imperative style, high-performance deep learning library.
\newblock {\em Advances in neural information processing systems}, 32, 2019.

\bibitem{radford2021clip}
Alec Radford, Jong~Wook Kim, Chris Hallacy, Aditya Ramesh, Gabriel Goh, Sandhini Agarwal, Girish Sastry, Amanda Askell, Pamela Mishkin, Jack Clark, et~al.
\newblock Learning transferable visual models from natural language supervision.
\newblock In {\em International conference on machine learning}, pages 8748--8763. PMLR, 2021.

\bibitem{romero2014fitnets}
Adriana Romero, Nicolas Ballas, Samira~Ebrahimi Kahou, Antoine Chassang, Carlo Gatta, and Yoshua Bengio.
\newblock Fitnets: Hints for thin deep nets.
\newblock {\em arXiv preprint arXiv:1412.6550}, 2014.

\bibitem{sultani2018real}
Waqas Sultani, Chen Chen, and Mubarak Shah.
\newblock Real-world anomaly detection in surveillance videos.
\newblock In {\em Proceedings of the IEEE conference on computer vision and pattern recognition}, pages 6479--6488, 2018.

\bibitem{tian2019contrastive}
Yonglong Tian, Dilip Krishnan, and Phillip Isola.
\newblock Contrastive representation distillation.
\newblock {\em arXiv preprint arXiv:1910.10699}, 2019.

\bibitem{tian2021weakly}
Yu Tian, Guansong Pang, Yuanhong Chen, Rajvinder Singh, Johan~W Verjans, and Gustavo Carneiro.
\newblock Weakly-supervised video anomaly detection with robust temporal feature magnitude learning.
\newblock In {\em Proceedings of the IEEE/CVF international conference on computer vision}, pages 4975--4986, 2021.

\bibitem{wang2019gods}
Jue Wang and Anoop Cherian.
\newblock Gods: Generalized one-class discriminative subspaces for anomaly detection.
\newblock In {\em Proceedings of the IEEE/CVF International Conference on Computer Vision}, pages 8201--8211, 2019.

\bibitem{wang2020cluster}
Ziming Wang, Yuexian Zou, and Zeming Zhang.
\newblock Cluster attention contrast for video anomaly detection.
\newblock In {\em Proceedings of the 28th ACM international conference on multimedia}, pages 2463--2471, 2020.

\bibitem{wu2022self}
Jhih-Ciang Wu, He-Yen Hsieh, Ding-Jie Chen, Chiou-Shann Fuh, and Tyng-Luh Liu.
\newblock Self-supervised sparse representation for video anomaly detection.
\newblock In {\em Computer Vision--ECCV 2022: 17th European Conference, Tel Aviv, Israel, October 23--27, 2022, Proceedings, Part XIII}, pages 729--745. Springer, 2022.

\bibitem{wu2021learning}
Peng Wu and Jing Liu.
\newblock Learning causal temporal relation and feature discrimination for anomaly detection.
\newblock {\em IEEE Transactions on Image Processing}, 30:3513--3527, 2021.

\bibitem{wu2020not}
Peng Wu, Jing Liu, Yujia Shi, Yujia Sun, Fangtao Shao, Zhaoyang Wu, and Zhiwei Yang.
\newblock Not only look, but also listen: Learning multimodal violence detection under weak supervision.
\newblock In {\em Computer Vision--ECCV 2020: 16th European Conference, Glasgow, UK, August 23--28, 2020, Proceedings, Part XXX 16}, pages 322--339. Springer, 2020.

\bibitem{xu2015learning}
Dan Xu, Elisa Ricci, Yan Yan, Jingkuan Song, and Nicu Sebe.
\newblock Learning deep representations of appearance and motion for anomalous event detection.
\newblock {\em arXiv preprint arXiv:1510.01553}, 2015.

\bibitem{yu2020cloze}
Guang Yu, Siqi Wang, Zhiping Cai, En Zhu, Chuanfu Xu, Jianping Yin, and Marius Kloft.
\newblock Cloze test helps: Effective video anomaly detection via learning to complete video events.
\newblock In {\em Proceedings of the 28th ACM International Conference on Multimedia}, pages 583--591, 2020.

\bibitem{zaheer2020claws}
Muhammad~Zaigham Zaheer, Arif Mahmood, Marcella Astrid, and Seung-Ik Lee.
\newblock Claws: Clustering assisted weakly supervised learning with normalcy suppression for anomalous event detection.
\newblock In {\em Computer Vision--ECCV 2020: 16th European Conference, Glasgow, UK, August 23--28, 2020, Proceedings, Part XXII 16}, pages 358--376. Springer, 2020.

\bibitem{zhang2018s3d}
Da Zhang, Xiyang Dai, Xin Wang, and Yuan-Fang Wang.
\newblock S3d: single shot multi-span detector via fully 3d convolutional networks.
\newblock {\em arXiv preprint arXiv:1807.08069}, 2018.

\bibitem{zhang2019temporal}
Jiangong Zhang, Laiyun Qing, and Jun Miao.
\newblock Temporal convolutional network with complementary inner bag loss for weakly supervised anomaly detection.
\newblock In {\em 2019 IEEE International Conference on Image Processing (ICIP)}, pages 4030--4034. IEEE, 2019.

\bibitem{zhang2020self}
Zhilu Zhang and Mert Sabuncu.
\newblock Self-distillation as instance-specific label smoothing.
\newblock {\em Advances in Neural Information Processing Systems}, 33:2184--2195, 2020.

\bibitem{zhong2019graph}
Jia-Xing Zhong, Nannan Li, Weijie Kong, Shan Liu, Thomas~H Li, and Ge Li.
\newblock Graph convolutional label noise cleaner: Train a plug-and-play action classifier for anomaly detection.
\newblock In {\em Proceedings of the IEEE/CVF conference on computer vision and pattern recognition}, pages 1237--1246, 2019.

\end{thebibliography}
}

\end{document}